\documentclass[journal]{IEEEtran}
\usepackage{amsmath,amsfonts}
\usepackage{algorithm}
\usepackage{array}
\usepackage{textcomp}
\usepackage{stfloats}
\usepackage{url}
\usepackage{verbatim}
\usepackage{graphicx}
\usepackage{cite}
\usepackage{subfigure}
\usepackage{multirow}
\usepackage{booktabs}
\usepackage{makecell}
\usepackage{amssymb}
\usepackage{arydshln}
\usepackage{algpseudocode}
\usepackage{algorithmicx}
\usepackage{color}
\newcounter{ToDo}
\newcounter{gaocomm}
\newcounter{wangcomm}
\newcounter{Note}
\definecolor{blue-violet}{rgb}{0.00,0.75,0.90}
\definecolor{mygreen}{rgb}{0.0, 0.5, 0.0}
\definecolor{awesome}{rgb}{1.0, 0.13, 0.32}
\definecolor{bostonuniversityred}{rgb}{1.0, 0.0, 0.0}

\begin{document}

\title{Hierarchical Multi-modal Transformer for Cross-modal Long Document Classification}

\author{Tengfei Liu,
        Yongli Hu, \IEEEmembership{Member,~IEEE},
        Junbin Gao,
        Yanfeng Sun, \IEEEmembership{Member,~IEEE},\\
        and Baocai Yin, \IEEEmembership{Member,~IEEE}
\thanks{Tengfei Liu, Yongli Hu, Yanfeng Sun and Baocai Yin are with Beijing Key Laboratory of Multimedia and Intelligent Software Technology, Beijing Institute of Artificial Intelligence, Faculty of Information Technology, Beijing University of Technology, Beijing 100124, China. E-mail: tfliu@emails.bjut.edu.cn, \{huyongli, yfsun, ybc\}@bjut.edu.cn (Corresponding author: Yongli Hu)}
\thanks{Junbin Gao is with the Discipline of Business Analytics, The University of Sydney Business School, The University of Sydney, Camperdown, NSW 2006, Australia. \protect E-mail: junbin.gao@sydney.edu.au}
}
\markboth{IEEE TRANSACTIONS ON MULTIMEDIA}%
{Liu \MakeLowercase{\textit{et al.}}: Hierarchical Multi-modal Transformer for Cross-modal Long Document Classification}

\maketitle

\begin{abstract}
Long Document Classification (LDC) has gained significant attention recently. However, multi-modal data in long documents such as texts and images are not being effectively utilized. Prior studies in this area have attempted to integrate texts and images in document-related tasks, but they have only focused on short text sequences and images of pages. How to classify long documents with hierarchical structure texts and embedding images is a new problem and faces multi-modal representation difficulties. In this paper, we propose a novel approach called Hierarchical Multi-modal Transformer (HMT) for cross-modal long document classification. The HMT conducts multi-modal feature interaction and fusion between images and texts in a hierarchical manner. Our approach uses a multi-modal transformer and a dynamic multi-scale multi-modal transformer to model the complex relationships between image features, and the section and sentence features. Furthermore, we introduce a new interaction strategy called the dynamic mask transfer module to integrate these two transformers by propagating features between them. To validate our approach, we conduct cross-modal LDC experiments on two newly created and two publicly available multi-modal long document datasets, and the results show that the proposed HMT outperforms state-of-the-art single-modality and multi-modality methods.
\end{abstract}

\begin{IEEEkeywords}
Cross-modal long document classification, multi-modal transformer, dynamic multi-scale multi-modal transformer, dynamic mask transfer.
\end{IEEEkeywords}

\section{Introduction}
\IEEEPARstart{A}{s} the number of accessible documents and publications grows exponentially, long document-related tasks have gained significant attention and made substantial progress. These tasks include long document summarization \cite{Cui2021SlidingSN}, long document question answering \cite{Nie2022CapturingGS}, long document machine reading comprehension \cite{Zhao2021RoRRF}, and long document classification (LDC) \cite{beltagy2020longformer, Liu2022HierarchicalGC}. As a fundamental task in Natural Language Processing (NLP), LDC plays an important role in many scenarios such as document management, document analysis, and personalized document recommendation \cite{bowen2009document, guan2010document}. Recent studies in LDC have focused on the development of novel methods and techniques to better address the challenges posed by long documents.

Transformer \cite{vaswani2017attention} has gained widespread popularity in various NLP tasks due to its exceptional performance \cite{Yang2019XLNetGA, Zaheer2020BigBT, Heo2022HypergraphTW}. However, the self-attention mechanism in Transformer has quadratic growth in memory usage and computational complexity with the sequence length. As a result, processing long documents with Transformer can be difficult and very expensive. To address this issue, two primary approaches have been proposed. One approach involves dividing long documents into manageable, isometric chunks using sequence cutting or sliding windows, followed by hierarchical modeling using Transformers. The other approach involves replacing the fully quadratic self-attention scheme of the Transformer with a sparse-attention mechanism. This modification enables the Transformer to handle documents with thousands of tokens or longer. However, it is important to note that these methods primarily focus on long text processing.

In recent years, researchers have made attempts to address multimodal document-related tasks such as visually rich document understanding \cite{Gu2022XYLayoutLMTL}, multimodal document quality assessment \cite{Shen2020AGA}, and document image classification \cite{Bakkali2022VLCDoCVC}. These attempts primarily focus on how to leverage and incorporate information from different modalities such as vision, language, and layout to obtain more effective multi-modal representations. However, most of these studies consider document pages as images and concentrate on the text stream extracted from these images using OCR \cite{Smith2007AnOO}, failing to explore the complex structure of long documents and the interaction between the texts and the embedding images.

\begin{figure}[t]
  \centering\includegraphics[width=1.0\linewidth]{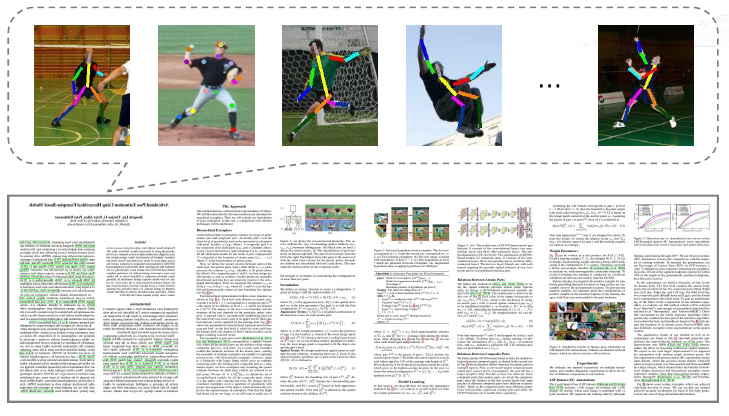}
   \caption{Example of multi-modal long document from the dataset of MAAPD.}
   \label{fig 0}
\end{figure}

\begin{figure}[t]
  \centering\includegraphics[width=1.0\linewidth]{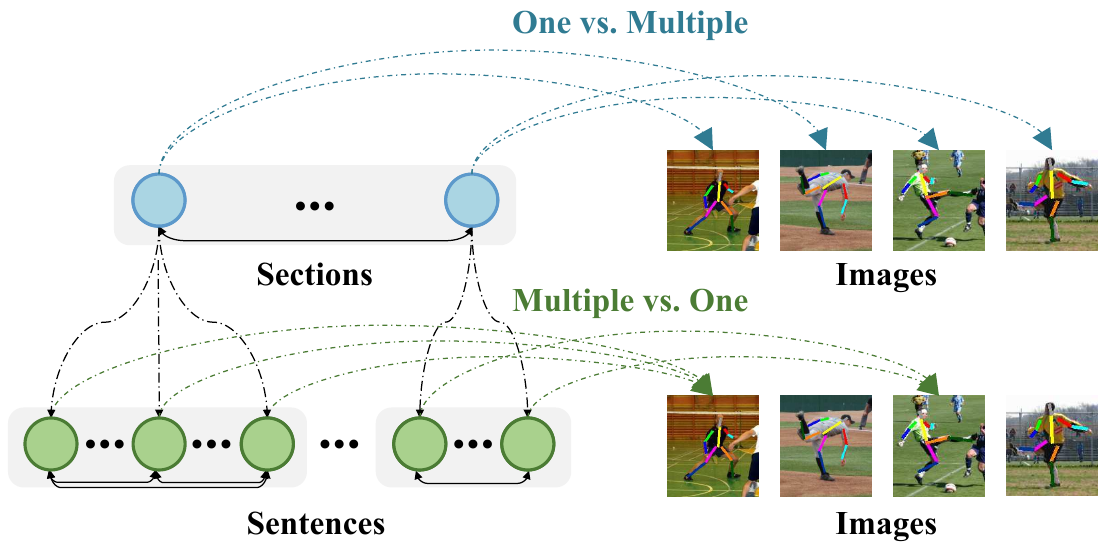}
   \caption{Diagram of the hierarchical text structure of long documents, and their corresponding relations with the paired images.}
   \label{fig 1}
\end{figure}

Long documents, such as scientific papers shown in Fig.~\ref{fig 0} and other publications, typically feature long text sequences with  explicit or implicit hierarchical structures and multiple embedding images. The challenge in cross-modal long document representation and classification is to fully exploit the hierarchical structure of the document and the complementary signals of visual and textual information. This problem differs from common multimodal tasks such as image-text retrieval \cite{Guo2022HGANHG}, multi-modal emotion analysis \cite{Nie2021CGCNCB}, or visual question answering \cite{Huasong2021SelfAdaptiveNM}. In those tasks,  the relations between images and texts are simple and definitive, and the text length is typically short. To understand the nature of cross-modal long documents, consider a scientific document, as illustrated in Fig.~\ref{fig 1}. 
First, the texts of the long document are organized in a hierarchical manner, i.e., word-sentence-section, associated with several images. Second, as shown in Fig.~\ref{fig 1}, the relationships between the images and the texts vary across different levels. 
For example, a section is typically described by multiple images, while an image is often described by a number of sentences. Lastly, there may be weak correlations between the images and the texts, and some images may even be unrelated, which adds difficulties and poses challenges to the multi-modal representation of long documents.

Based on the previous analysis, we introduce a new approach called Hierarchical Multi-modal Transformer (HMT) for Cross-modal Long Document Classification (CLDC). This framework is meticulously designed to address the intricate characteristics of multimodal long documents. Specifically, to effectively model the structured information inherent in long documents, we construct section-level and sentence-level features separately. This dual-level feature construction not only accurately captures the hierarchical structure of long documents but also facilitates the subsequent multi-granularity relationship capture between textual and visual modalities. For multimodal modeling, we employ state-of-the-art multimodal Transformers as our primary encoders. This architecture excels in performance and adeptly manages the weak correlations that often exist between text and images, thereby mitigating the potential impact of visual information on textual content. To fully leverage the hierarchical structure of the text, we introduce a dynamic mask transfer module. This innovative module ensures seamless information interaction between section-level and sentence-level multimodal Transformers, enabling higher-level structures to inform and enhance the processing at lower levels. These integrated modules synergistically work to organize and represent information in a hierarchical manner, effectively capturing complementary data from various modalities and maintaining a consistent and comprehensive flow of information. 

Fig.~\ref{fig 2} illustrates the detailed HMT architecture, which starts by utilizing pre-trained text and image models to capture the hierarchical text features, i.e., section and sentence features, and image features. Following this, two multi-modal transformers are employed to capture the complex relationships between the textual and visual features at different levels. These transformers are structured to handle hierarchical data, ensuring that both section-level and sentence-level interactions are effectively modeled. Specifically, to model the multi-scale relations between sentences and images, we introduce an enhanced transformer, namely Dynamic Multi-scale Multi-modal Transformer (DMMT). The DMMT is designed by assigning window masks of different sizes to the multi-head attention matrix, which allows it to capture varying levels of detail in the interactions. The multi-scale window branches are then dynamically fused with different weights, ensuring that the most relevant features are highlighted in the final representation. Finally, a dynamic mask transfer module is incorporated to fully leverage hierarchical text associations. This module enables effective interaction between the two multi-modal transformers, allowing insights from one level (e.g., section-level) to inform and enhance processing at another level (e.g., sentence-level). Considering that most of the existing multi-modal long document datasets only contain text sequences with less than 500 tokens and insufficient images, we construct two new multi-modal long document datasets in the field of scientific papers, which will be made available publicly to comprehensively test the proposed method. We summarize our contributions as follows,

\begin{itemize}
    \item The proposed Hierarchical Multi-modal Transformer (HMT) integrates text and image features at various levels of granularity for Cross-modal Long Document Classification (CLDC). To the best of our knowledge, this approach is the first of its kind to consider both the hierarchical structure and visual image information present in long documents for the task of CLDC.
    
    \item Two well-designed techniques, namely  Dynamic Multi-scale Multi-modal Transformer (DMMT) and Dynamic Mask Transfer (DMT) module, are proposed to effectively capture the complex relationships between sentences and images in long documents and enhance the interaction between the two multi-modal transformers.
    
    \item Results from experiments conducted on two newly created and two publicly available multi-modal long document datasets indicate that the proposed method outperforms single-modal text methods and surpasses state-of-the-art multi-modal baselines. 
\end{itemize}

The remainder of the paper is organized as follows. Related works are surveyed in Section~\ref{related work}, and the HMT framework is presented in Section~\ref{The Proposed Model}. The evaluation is conducted in Section~\ref{Experiments} and the paper is concluded in Section~\ref{Conclusion}.

\section{Related Work}\label{related work}
\subsection{Long Document Classification}
The task of LDC involves dealing with long documents  with explicit or implicit hierarchical structures, and usually, their text contents can span thousands of tokens. Traditional document classification techniques may not be effective in such cases, as they are not capable of capturing long-range dependencies. The existing works for long document classification can be divided into two main categories. 

The first category of methods for long document classification involves representing long documents in a hierarchical manner. For example, Pappagari et al. \cite{pappagari2019hierarchical} divided the long text into manageable segments and fed each of them into the base model of BERT \cite{devlin2018bert}. The inter-segment interactions are then modeled by applying a single recurrent layer or transformer to the segment representations. Similarly, Wu et al. \cite{Wu2021HiTransformerHI} proposed a hierarchical interactive transformer, which models documents in a hierarchical manner and can capture the global document context for sentence modeling. Likewise, Liu et al. \cite{Liu2022HierarchicalGC} proposed a hierarchical graph convolutional network for LDC, where a section graph and a word graph are constructed to model the macro and microstructure of long documents, respectively.  

An alternative category of methods aims to address the computational complexity of the self-attention mechanism of the Transformer \cite{vaswani2017attention}, which limits its ability to process long texts of thousands of tokens. So far, different sparsity mechanisms have been explored, e.g., the fixed pattern \cite{beltagy2020longformer, Zaheer2020BigBT}, the learnable pattern \cite{Piao2022SparseSL}, and the low-rank pattern \cite{Wang2020LinformerSW}. 

In the case of fixed pattern sparsity mechanisms, the aim is to limit the scope of attention. This is achieved through window-based attention, global attention, and random attention techniques. For instance, Beltagy et al. \cite{beltagy2020longformer} proposed a combination of windowed local-context attention and task-motivated global attention, successfully reducing the complexity from quadratic to linear. Zaheer et al. \cite{Zaheer2020BigBT} further integrated random attention into the model proposed in \cite{beltagy2020longformer} and also achieved linear complexity with respect to the number of tokens. The learnable pattern tries to capture both local and global contexts more effectively. For example, Kitaev et al. \cite{kitaev2020reformer} introduced a learnable pattern using locality-sensitive hashing to identify the nearest neighbors of the attention query for attention. Zhang et al. \cite{Zhang2021PoolingformerLD} proposed a two-level attention schema that combines sliding window and pooling attention to reduce computational and memory costs. The low-rank pattern \cite{Wang2020LinformerSW} observes that the self-attention matrices are low-rank. Thus, it chooses to linearly project key and value matrices into a low-dimensional space, e.g., from $n$ to $k$, to achieve a $\mathcal{O}(nk)$ complexity.

Despite clear advantages, these methods mainly focus on modeling text of long documents, while the cross-modal information is not considered.

\begin{figure*}[t]
  \centering
   \includegraphics[width=0.88 \linewidth]{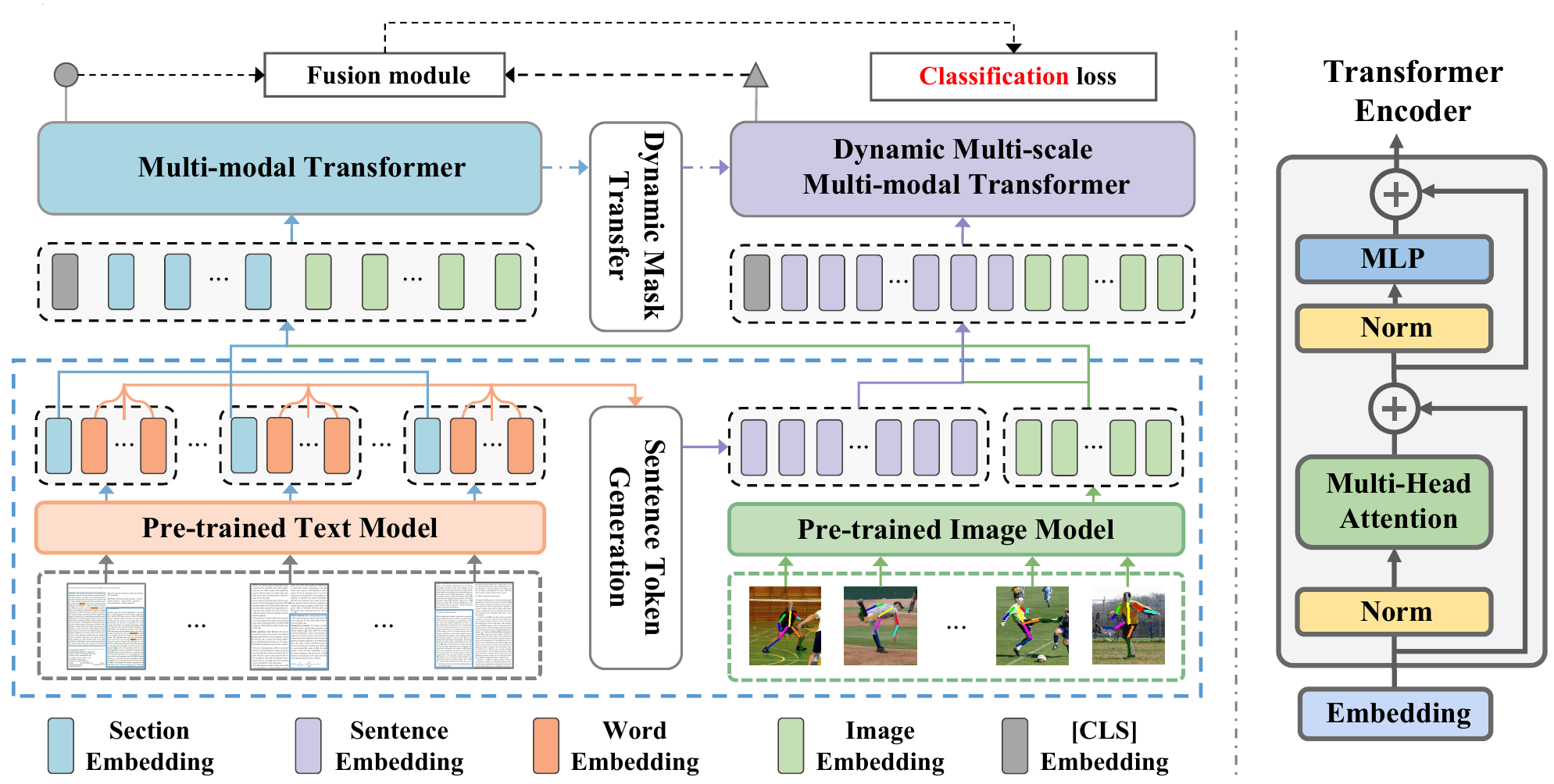}
   \caption{The framework of the proposed HMT model for cross-modal long document classification.}
   \label{fig 2}
\end{figure*}

\subsection{Document Image Classification}
Considerable research has been conducted to effectively organize and index document images. These approaches \cite{Ferrando2020ImprovingAA, Bakkali2022VLCDoCVC} typically involve two streams of information: images and texts. In this pipeline, images serve as the visual representation of document pages, while the text is usually obtained by applying OCR \cite{Smith2007AnOO} to the images. The rapid progress of deep learning has driven further advancements in this domain, enabling the automatic extraction of textual and visual features from document images \cite{Asim2019TwoSD}. While document image classification methods have made significant progress in recent years, they often overlook the real images within documents and do not take full advantage of the hierarchical structure of the text. As a result, there is still room for improvement in these methods. For this purpose, we propose a novel hierarchical multi-modal transformer for CLDC, in which the hierarchical structure of texts and the embedding images of long documents are utilized, instead of the page images. 

\subsection{Multi-modal Transformer}
The multi-modal transformer has proven to be effective in various multi-modal tasks, including image/video-text retrieval \cite{Guo2022HGANHG,wang2022align}, visual question answering \cite{Huasong2021SelfAdaptiveNM}, multi-modal emotional analysis \cite{Nie2021CGCNCB}, and visual grounding \cite{wang2021weakly}, 
resulting in impressive performance outcomes.

Currently, there are two main approaches for incorporating image data into a multi-modal transformer for text-image tasks. The first approach involves dividing the image into several patches, which are then treated as separate image inputs to the model. The second approach uses object detection techniques, such as Faster-RCNN \cite{Li2022ActionAwareEE}, to extract region features from the image, and considers them as the input to the model. In some scenarios, to improve the cross-modal representation capability, additional data may also be incorporated. For instance, scene text embeddings extracted from the image \cite{cheng2022vista} and class labels \cite{lin2022cross} can be utilized. In the context of text-video data, most existing approaches treat each image frame as a single feature, which is then used to interact with the text features. Our image-processing technique is similar to those used in text-video-related tasks, but with some notable differences. Firstly, our images are unordered. Additionally, the interaction objects for the images are not fine-grained word features, but high-level semantic features such as section and sentence features. 

While the multi-modal transformer has proven to be effective in a range of multi-modal tasks, it has not yet been applied to the CLDC task. To address this, we introduce the HMT model, which incorporates a multi-modal transformer and a dynamic multi-scale multi-modal transformer to capture the intricate relationships between image features and hierarchical text features such as section and sentence features. By leveraging these advanced modeling techniques, we aim to improve the performance of CLDC and push the boundaries of document understanding.

\section{Proposed Method}\label{The Proposed Model}
In this section, we present the proposed Hierarchical Multi-modal Transformer (HMT) in detail. As shown in Fig.~\ref{fig 2}, the pre-trained text and image models are first employed to obtain initial hierarchical text features and image features. Next, the complex multi-modal relationships between these features are modeled using a hierarchical multi-modal transformer at both the section and sentence levels. Additionally, we introduce a dynamic mask transfer module to facilitate information exchange between the two multi-modal transformers. Finally, a fusion module is applied to aggregate the final multi-modal representations at both levels, 
enabling the classification of long documents.

\subsection{Feature Extraction}
\subsubsection{Textual Features} 
To model the complex association of image features and text features at the section and sentence levels, we segment each long document $t$ into $l$ sequential sections in a fixed size without overlapping, denoted as $\{q_{1}, q_{2},...,q_{l}\}$. Each section contains $r$ word tokens, i.e., $q_{i} = \{w_{i0},w_{i1},...,w_{ir}\}$, where $w_{i0}$ = $\texttt{[CLS]}$ is the special start token for BERT-related encoders. Extra $\texttt{[PAD]}$ tokens are appended to the end to meet the section length of $r$. Then, we apply the pre-trained BERT \cite{devlin2018bert} $f_{t}(\cdot,\phi)$, which has been widely utilized in LDC tasks \cite{pappagari2019hierarchical, Liu2022HierarchicalGC}, to extract the $i$th section feature with word features as follows,
\begin{equation}
     [p_{i}, x_{i1},...,x_{ir}]=f_{t}(q_i,\phi), i=1,...,l.\label{fomulate 1}
\end{equation}
where $\phi$ represents the parameters of the pre-trained BERT. The final hidden state of the $\texttt{[CLS]}$ token, $p_{i}$, is taken as the section feature, and the other tokens $x_{ij},j=1,...,r$ are taken as the word features. The final section features and word features of the long document can be represented as matrix features $P = [p_1,p_2,...,p_l] \in \mathbb{R}^{l \times d}$ and  $X = [x_{11},...,x_{1r},...,x_{l1},...,x_{lr}] \in \mathbb{R}^{lr \times d}$, where $d$ is the dimension of the features. 

\begin{figure}[t]
  \centering
   \includegraphics[width=1.0\linewidth]{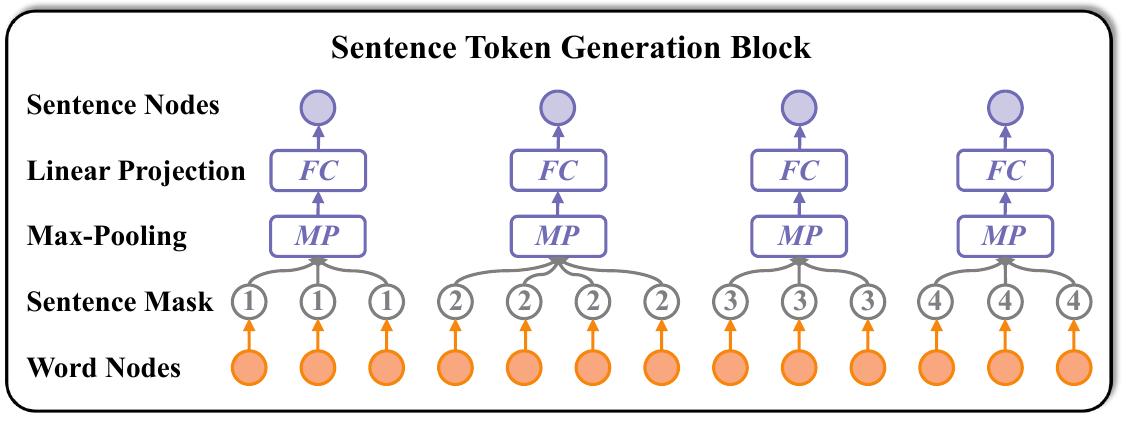}
   \caption{An illustration of the proposed Sentence Token Generation (STG) block.}
   \label{fig 3}
\end{figure}

To make the most of the fine-grained word features and avoid additional data input, the sentence features are directly obtained by aggregating the word features in each sentence by the proposed Sentence Token Generation (STG) block as shown in Fig.~\ref{fig 3}. This process is directed by the sentence mask $S_{mask} \in \mathbb{R}^{lr}$ generated from the document. It is represented by an incrementally repeatable number sequence, in which the mask values of words in the same sentence keep fixed. The final sentence features are obtained by applying the max-pooling operation to the word features in each sentence, followed by a linear projection layer as follows, 
\begin{equation}
\begin{gathered}
     \bar{s}_{i}=\text{maxpooling\_c}(X[S_{mask} == i\textbf{1}_{lr}]), i=1,...,n.\\
     s_{i}=W_{s}\bar{s}_{i} + b_{s}, i=1,...,n.\label{fomulate 2}
\end{gathered}
\end{equation}
where $\text{maxpooling\_c}$ and $n$ represent the column-wise max-pooling and the number of sentences, respectively. $\textbf{1}_{lr} \in \mathbb{R}^{lr}$ is a vector in the size of $lr$ with all elements being 1. $W_{s}$, $b_{s}$ are the trainable parameters of the linear projection layer. The final sentence features of the long document can be denoted as $S = [s_1,s_2,...,s_n] \in \mathbb{R}^{n \times d}$. 

\subsubsection{Visual Features} 
Different from the traditional multi-modal tasks, which focus on the local region features of images, we pay more attention to the global semantic representations for images $v_{j},j=1,2,...,m$, where $m$ is the number of images, in long documents, considering their better correspondences with the section features and sentence features. In light of the success of vision transformer (ViT) \cite{Dosovitskiy2021AnII}, which takes non-overlapping image patches as the input, we directly utilize the pre-trained CLIP model \cite{Radford2021LearningTV} with a ViT architecture as our image encoder $f_{v}(\cdot,\psi)$. The CLIP model performs self-attention between patches and compiles the set of patch embeddings into a $\texttt{[CLS]}$ embedding. We then leverage a fully connected layer to project all the image embeddings to the same dimension as the text embeddings, as follows,
\begin{equation}
     h_{j}=W_{h}(f_{v}(v_{j},\psi)) + b_{h}, j=1,...,m.\label{fomulate 3}
\end{equation}
where $\psi$ represents the parameters of the pre-trained image encoder. The final image features of the long document can be formulated as $H = [h_1,h_2,...,h_m] \in \mathbb{R}^{m \times d}$. $d$ is the dimension of the features.

\subsection{Hierarchical Multi-modal Transformer}
As illustrated in Fig.~\ref{fig 1}, the text and image features of a long document have varying degrees of interdependence at different hierarchical levels, such as one-to-multiple and multiple-to-one relationships. To capture these complex and diverse relationships, we propose two multi-modal transformers, namely the Multi-modal Transformer and the Dynamic Multi-scale Multi-modal Transformer. Furthermore, we introduce a dynamic mask transfer module to enable information interaction between these two levels of multi-modal transformers.

\subsubsection{Multi-modal Transformer}
To model the one-to-multiple relationships between sections and images, we directly concatenate the $\texttt{[CLS]}$ token, section features $P$ and image features $H$ into a unified multi-modal sequence $F_{pv} = [\texttt{[CLS]}, p_1,p_2,...,p_l, h_1,h_2,...,h_m]$. Then, following the same architecture of the Transformer \cite{vaswani2017attention}, we build our multi-modal transformer with a stack of multi-head self-attention layers followed by a feed-forward network, which can be denoted as follows,
\begin{equation}
\begin{gathered}
    E_{pv} \leftarrow F_{pv} + \text{MHSA}(\text{LN}(F_{pv})),\\
    \hat{F}_{pv} \leftarrow E_{pv} + \text{MLP}(\text{LN}(E_{pv}))\label{fomulate 4}
\end{gathered}
\end{equation}
where $\text{MHSA}(\cdot)$ denotes the multi-head self-attention layer, $\text{MLP}(\cdot)$ denotes the multi-layer perception layer, and $\text{LN}(\cdot)$ denotes the layer normalization. The final output of the $\texttt{[CLS]}$ token of $\hat{F}_{pv}$ is adopted as the fused multi-modal representation in the section level, denoted as $y_{pv}$.

\begin{figure}[t]
  \centering
   \includegraphics[width=\linewidth]{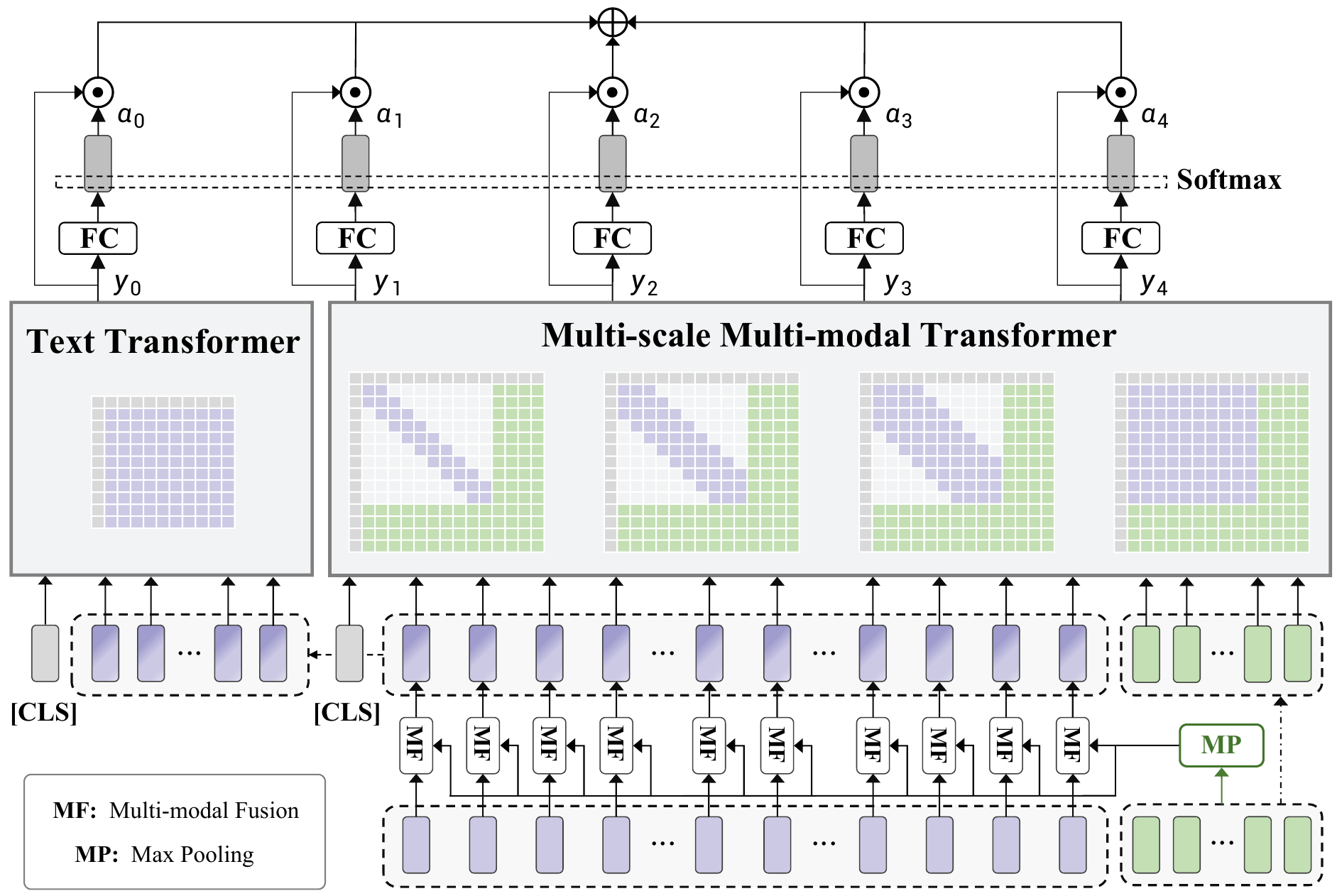}
   \caption{Schematic diagram of the Dynamic Multi-scale Multi-modal Transformer (DMMT).}
   \label{fig 4}
\end{figure}

\subsubsection{Dynamic Multi-scale Multi-modal Transformer}
Fig.~\ref{fig 4} illustrates the proposed Dynamic Multi-scale Multi-modal Transformer (DMMT), which consists of two branches: a single text transformer and a multi-scale multi-modal transformer. The former is designed to mitigate the negative impact of uncorrelated images on the final output, while the latter is responsible for capturing the multi-scale corresponding relationships between the sentence and image features.

To reduce the heterogeneous gap and enhance the representation power of multi-modal features, we first adopt a simple fusion strategy to generate multi-modal sentence representations. Specifically, given the max-pooling feature $I$ of the image features and the sentence features $[s_{1},s_{2},...,s_{n}]$,  the procedure is formulated as follows,
\begin{equation}
\begin{gathered}
    I = \text{maxpooling\_c}(H)\\
    \hat{s}_{i} = W_{c}[s_{i};I] + b_{c},i=1,2,...,n.\label{fomulate 5}
\end{gathered}
\end{equation}
where $\text{maxpooling\_c}$ represents the column-wise max-pooling. $[\cdot;\cdot]$ denotes concatenation along the feature dimension. $W_{c} \in \mathbb{R}^{2d \times d}$ and $b_{c} \in \mathbb{R}^{d}$ are the trainable parameters. The fused multi-modal sentence features can be represented as $\hat{S} = [\hat{s}_{1},\hat{s}_{2},...,\hat{s}_{n}]$. 

Fig.~\ref{fig 1} demonstrates that an image is usually depicted by several different sentences. To model the multi-scale correspondence relationships between sentences and images, we choose to limit the scope of attention of sentence elements by placing multi-scale window masks on the multi-head adjacency matrix. Concretely, given the multi-modal input sequence $F_{sv} = [\texttt{[CLS]}, \hat{S}, H]$, and the window mask matrix $\text{D} \in \mathbb{R}^{(n+m+1) \times (n+m+1)}$ under one of the scales, the above process can be formulated as,
\begin{equation}
\begin{aligned}
     \vec{F}_{sv} &= \text{MHSA}(\bar{F}_{sv})\\
                  &= W_{o}\text{Concat}(A_{1}V_{1},A_{1}V_{2},...,A_{h}V_{h})\\
    A_{i} &= \text{Softmax}(\frac{(\bar{F}_{sv}W_{i}^{q})(\bar{F}_{sv}W_{i}^{k})^{T}}{\sqrt{d}} \odot \text{D})\label{fomulate 6}
\end{aligned}
\end{equation}
where $i=1,2,...,h$ and the values of $\text{D}$ are binary and set to 1 if they are within the attention span of the target element. $\text{MHSA}(\cdot)$ denotes the multi-head self-attention block. $\bar{F}_{sv} = \text{LN}(F_{sv})$, $V_{i} = \bar{F}_{sv}W_{i}^{v}$ and $W_{i}^{q} \in \mathbb{R}^{d \times d_{k}}, W_{i}^{k} \in \mathbb{R}^{d \times d_{k}}, W_{i}^{v} \in \mathbb{R}^{d \times d_{v}}, W_{o} \in \mathbb{R}^{d \times d}$ are the trainable parameters of MHSA. Especially, $d_{k}=d_{v}=d/h$ and $h$ is the total number of heads in the MHSA block. Fig.~\ref{fig 4} presents an architecture diagram of DMMT with $n_{win} = 4$ scale window mask matrices. It is noted that a fully-connected attention mask is included to capture the global information interaction between sentences and images. The text branch follows the same learning rules of Transformer \cite{vaswani2017attention}, and the multi-modal branch owes the same learning procedure, but with a different MHSA mechanism as  \eqref{fomulate 6}, which are denoted by $\text{T-Transformer}(\cdot),\text{M-Transformer}(\cdot)$, respectively and formulated in \eqref{fomulate 7} as follows,
\begin{equation}
\begin{gathered}
    y_{0} = \text{T-Transformer}([\texttt{[CLS]},\hat{S}])\\
    y_{i} = \text{M-Transformer}(F_{sv}, \text{D}_{i}),i=1,2,...,n_{win}.\label{fomulate 7}
\end{gathered}
\end{equation}
where $y_{i},i=0,1,,...,n_{win}$ is the final output of the \texttt{[CLS]} tokens, and $\text{D}_i$ represents the $i$th window mask matrix. To this end, we design a dynamic multi-scale multi-modal information weighting strategy, which integrates information of all branches by the generated weights. It mainly consists of two fully-connected layers $F_{1}$, $F_{2}$, and an activation layer. Finally, a softmax layer is adopted to generate different weights for corresponding elements of all branches. The calculation process can be formulated as follows, 
\begin{equation}
\begin{gathered}
    y_{fuse} = F_{2}(\delta(F_{1}(y)))\\
    \alpha_{i} = \frac{e^{y_{fuse}^{i}}}{\sum_{i=0}^{n_{win}}e^{y_{fuse}^{i}}},i=0,1,,...,n_{win}.\label{fomulate 8}
\end{gathered}
\end{equation}
where $y_{fuse} \in \mathbb{R}^{(n_{win}+1) \times d}$, and $\alpha_{i} \in \mathbb{R}^{1 \times d}$. The output of the DMMT module is given by,
\begin{equation}
    y_{sv} = \sum_{i=0}^{n_{win}}\alpha_{i} \odot y_{i}\label{fomulate 9}
\end{equation}

\begin{figure}[t]
  \centering
   \includegraphics[width=1.0\linewidth]{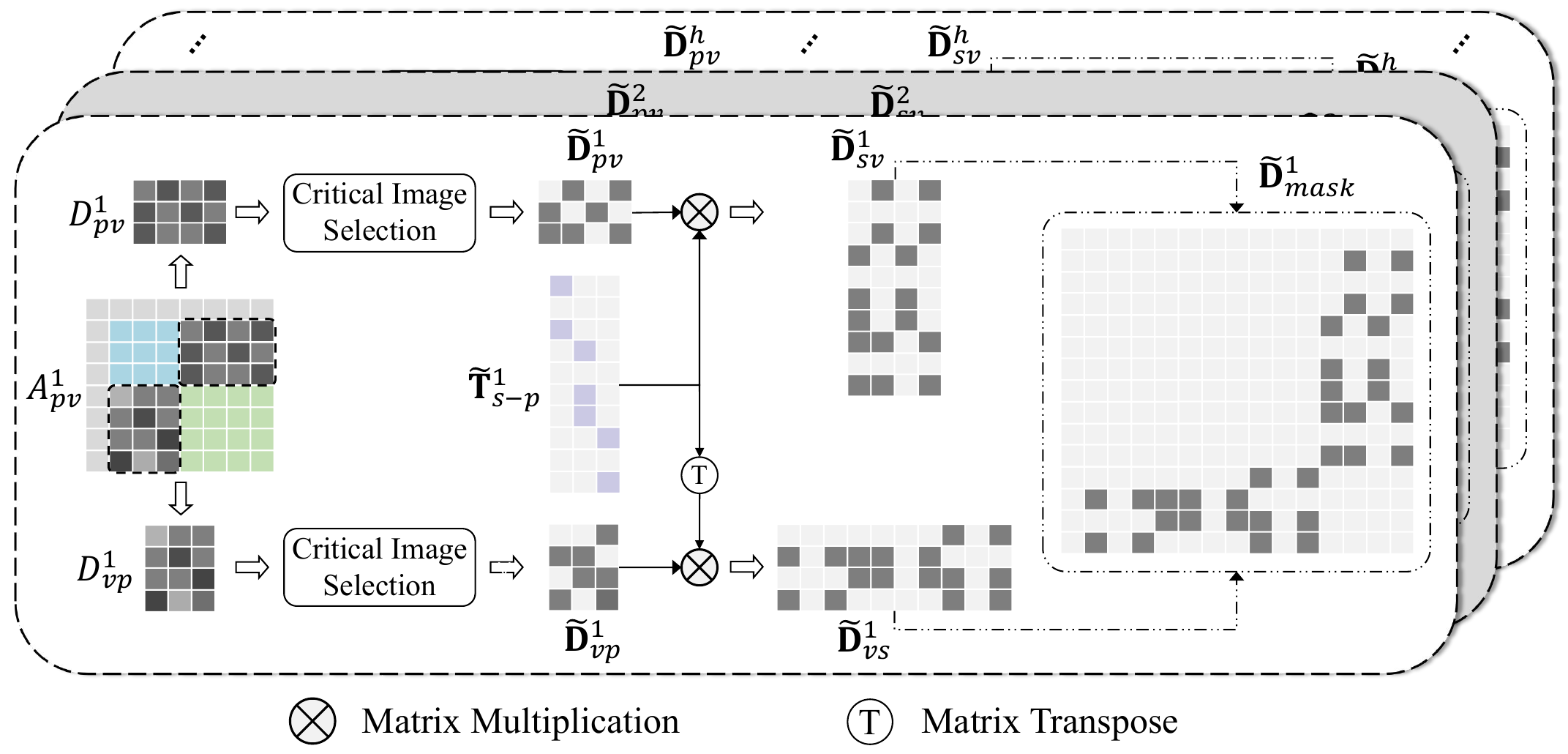}
   \caption{An illustration of the proposed Dynamic Mask Transfer (DMT) block.}
   \label{fig 5}
\end{figure}

\subsubsection{Dynamic Mask Transfer} 
In order to fully leverage the hierarchical structure of long documents, we apply the principle of transferability to facilitate the migration of information from the section-image association to the sentence-image association, thereby enhancing the cross-modal interaction capability between sentence and image features.

As shown in Fig.~\ref{fig 5}, we first extract the multi-head section-image adjacency matrix $D_{pv} \in \mathbb{R}^{h \times l \times m}$ and the multi-head image-section adjacency matrix $D_{vp} \in \mathbb{R}^{h \times m\times l}$ from the multi-head adjacency matrix $A_{pv} \in \mathbb{R}^{h \times (l+m+1) \times (l+m+1)}$ of Multi-modal Transformer.  Since the operations keep the same, here we mainly take the first head of $D_{pv}$ as an example to give the explanation.

For each section, we select the top-$\text{K}$ images with the highest similarity scores as the critical image index set $I^{+}$. The $\text{K}$ is a dynamic number for each section, which is the smallest number that meets the condition in \eqref{fomulate 10},
\begin{equation}
    \sum_{j \in I_{i}^{+}}\exp(D_{pv}^{1}(i,j))/\sum_{j=1}^{m}\exp(D_{pv}^{1}(i,j)) > \eta, i=1,2,...,l\label{fomulate 10}
\end{equation}
where $\eta$ is a constant, and we set $\eta=0.65$ in all experiments. 

According to the critical image index set $I_{i}^{+},i=1,2,...,l$, we can obtain the binary mask matrix $\tilde{\text{D}}_{pv}^{1}$ as follows,
\begin{equation}
 \tilde{\text{D}}_{pv}^{1}(i,j) = \begin{cases}
  1, j \in I_{i}^{+}\\
  0, \text{else}\label{fomulate 11}
\end{cases}
\end{equation}
where $i=1,2,...,l; j=1,2,...,m$. Then, we construct the transfer mask to model the relations between sentences and sections, denoted by $\mathbf{T}_{s-p} \in \mathbb{R}^{n \times l}$, in which if a sentence belongs to a section, the mask value will be set to 1, otherwise to 0, as follows, 
\begin{equation}
[\mathbf{T}_{s-p}]_{ij} = \begin{cases}
  1, \text{if the sentence node $i$ belongs to the section $j$}\\
  0, \text{else}\label{fomulate 12}
\end{cases}
\end{equation}

If we directly utilize $\mathbf{T}_{s-p} \times  \tilde{\text{D}}_{pv}^{1}$ , where $\times$ denotes matrix multiplication, to finish the mask transfer, it will inevitably bring a lot of noise. This is mainly because the section features occasionally have completely different semantics from the included sentence features, meaning that the images associated with section features are not definitely related to the included sentence features. To make the obtained sentence-image mask matrix more discriminative, we further sparse the transfer mask $\mathbf{T}_{s-p}$, only retaining  the sentence node masks that are relevant to the corresponding section features. This process is accomplished by placing a threshold on the cosine similarity matrix between the sentence features and corresponding section features, as follows,
\begin{equation}
\begin{gathered}
\mathbf{M}_{s-p}=\text{cos-similarity}(S,\mathbf{T}_{s-p}P)>0\\
\tilde{\mathbf{T}}_{s-p} = \mathbf{M}_{s-p} \times \mathbf{T}_{s-p}\label{fomulate 13}
\end{gathered}
\end{equation}
where $\mathbf{M}_{s-p} \in \mathbb{R}^{n \times 1}$. Given the sparse section-image mask matrix $\tilde{\text{D}}_{pv}^{1} \in \mathbb{R}^{l \times m}$ and the sparse transfer mask $\tilde{\mathbf{T}}_{s-p} \in \mathbb{R}^{n \times l}$,  the final sentence-image mask matrix can be formulated as,
\begin{equation}
\tilde{\text{D}}_{sv}^{1}=\tilde{\mathbf{T}}_{s-p} \times \tilde{\text{D}}_{pv}^{1}\label{fomulate 14}
\end{equation}

The same calculation procedures are applied to each head of the multi-head section-image attention matrix $D_{pv}$ and the multi-head image-section attention matrix $D_{vp}$. 
The final outputs can be expressed as $\tilde{\text{D}}_{sv}^{i} \in \mathbb{R}^{n \times m}, i=1,2,...,h$, and $\tilde{\text{D}}_{vs}^{i} \in \mathbb{R}^{m \times n}, i=1,2,...,h$, respectively. Then, as shown in Fig.~\ref{fig 5}, we reorganize them as a whole multi-head mask matrix $\tilde{\text{D}}_{mask}^{i} \in \mathbb{R}^{(n +m+1) \times (n+m+1)},i=1,2,...,h$ and integrate it into \eqref{fomulate 6} to enhance the interaction ability of sentence and image features, as follows,
\begin{equation}
\begin{aligned}
     \vec{F}_{sv} &= \text{MHSA}(\bar{F}_{sv})\\
                  &= W_{o}\text{Concat}(\hat{A}_{1}V_{1},\hat{A}_{1}V_{2},...,\hat{A}_{h}V_{h})\\
    \hat{A}_{i} &= \text{Softmax}((\frac{(\bar{F}_{sv}W_{i}^{q})(\bar{F}_{sv}W_{i}^{k})^{T}}{\sqrt{d}} \odot \text{D} \\ &\odot (1 + \tilde{\text{D}}_{mask}^{i})) \odot W_{m})\label{fomulate 15}
\end{aligned}
\end{equation}
where $i=1,2,...,h$ and 
 the dynamic weight $W_{m} \in \mathbb{R}^{(n+m+1) \times (n+m+1)}$ is added to maximize the contribution of the transfer mask $\tilde{\text{D}}_{mask}$ and minimize the influence of noise mask nodes on the multi-modal interaction at the sentence level. Accordingly, \eqref{fomulate 7} can be changed as follows,
\begin{equation}
\begin{gathered}
    y_{0} = \text{T-Transformer}([\texttt{[CLS]},\hat{S}])\\
    \hat{y}_{i} = \text{M-Transformer}(F_{sv}, \tilde{\text{D}}_{mask},\text{D}_{i}),i=1,2,...,n_{win}\label{fomulate 16}
\end{gathered}
\end{equation}

Finally, based on \eqref{fomulate 8} and \eqref{fomulate 9}, we can obtain the enhanced multi-modal representation $\hat{y}_{sv}$ at the sentence level.

\subsection{Model Training}
To obtain the final representation of the cross-modal long document, a column-wise max-pooling operation, denoted by maxpooling\_c($\cdot$), is applied on the multi-modal representations at the section and sentence levels as follows,
\begin{equation}
\begin{gathered}
     u = \text{maxpooling\_c}([y_{pv},\hat{y}_{sv}])\label{fomulate 17}
\end{gathered}
\end{equation}
where $[\cdot, \cdot]$ represents the concatenation operation along the column dimension. Then, we apply the softmax function to output the probability of each label and adopt the cross-entropy loss function as the classification loss as follows,
\begin{equation}
\begin{gathered}
    y_{c} = \text{softmax}(W_uu+b_u)\\
     \mathcal{L}_{cls} = -\sum_{i \in \mathcal{Y_{D}}} \sum_{j=1}^{F}Y_{ij}{\rm ln}y_{c,ij}\label{fomulate 18}
\end{gathered}
\end{equation}
where $\mathcal{Y_{D}}$ is the set of document indices referring to their labels, and $F$ is the dimension of the output feature, which is equal to the number of classes. $Y$ is the label indicator matrix. 

\begin{table*}
\renewcommand{\arraystretch}{1.4}
\caption{The statistical information of the multi-modal long document datasets used in this paper, in which the MMaterials and MAAPD are constructed by us.}
\label{table 1}
\centering
\setlength{\tabcolsep}{3.8mm}
\begin{tabular}{lccccccccccc}
\toprule
Datasets & \#Doc & \#Train & \#Val & \#Test & Avg.\#Sen & Avg.\#Word & \#Class & Avg.\#Img & Classification type\\
\midrule 
MMaterials & 9320 & 6812 & 1247 & 1261 & 206.6 & 4459.1 & 7 & 9 & Single-label\\
MAAPD & 24573 & 20753 & 2000 & 2000 & 301.5 & 6257.1  & 10 & 13 & Single-label\\
Review & 12940 & 10352 & 1294 & 1294 & 27.8 & 419.9  & 5 &  4.1 & Single-label\\
Food101 & 86102 & 58131 & 6452 & 21516 & 115.6 & 1930 & 101 & 1 & Single-label\\
\bottomrule
\end{tabular}
\end{table*}

\section{Experiments}\label{Experiments}
In this section, we evaluate the proposed HMT model on two newly created and two publicly available multi-modal long document datasets and compare it with the state-of-the-art LDC methods, including both single text and multi-modal methods. 

\subsection{Datasets}
In the field of visually-rich document classification, several datasets have been developed, such as RVL-CDP \cite{Harley2015EvaluationOD} and Tobacco-3482 \cite{Kumar2014StructuralSF}. However, these datasets only contain images of document pages without the internal embedding images. On the other hand, the commonly used cross-modal document classification datasets typically consist of short text sequences, which are not suitable for evaluating the CLDC task's capabilities. To address this issue, we have constructed two new multi-modal long document datasets, namely MMaterials, and MAAPD, in addition to two public datasets that are relatively suitable for our task. These datasets contain structured long text sequences and multiple embedding images obtained by the Grobid\footnote{\url{https://github.com/kermitt2/grobid}} and Fitz library, respectively. The specific statistics
of the four datasets are shown in Table~\ref{table 1}.
\begin{itemize}
\setlength{\itemsep}{3pt}
\setlength{\parsep}{0pt}
\setlength{\parskip}{0pt}
     \item \textbf{MMaterials}: We select 7 materials fields, i.e., composite material, battery separator, energy storage material, graphene, nanomaterial, silicon carbide, and titanium alloy, and download 9320 articles from the Internet. There are 6812 training samples, 1247 validation samples, and 1261 testing samples. Each sample has 4459 words and 9 images on average.
    
    \item \textbf{MAAPD}: We expand the dataset AAPD \cite{yang2018sgm} to a multi-modal version, i.e., MAAPD, and use it in the CLDC task. As shown in Table~\ref{table 1}, 24573 samples are obtained by automatically parsing the XML files and extracting the image information. The dataset contains 20753 training samples, 2000 validation samples, and 2000 testing samples, which are assigned to one of 10 subject categories such as \texttt{cs.cv}, \texttt{cs.cr}, \texttt{cs.ro}. Each document includes an average of 13.5 images.

     \item \textbf{Review}: The original online review dataset \cite{Truong2019VistaNetVA} has more than 44 thousand reviews, including 244 thousand images. However, the average text length of the samples is only 237.3 tokens. To better verify the effectiveness of our proposed model in processing long documents, we select the samples with text lengths larger than 256 and reorganize them as a new dataset, which contains 10352 training, 1294 validation, and 1294 testing samples with an average of 419.9 words, 27.8 sentences, and 4.1 images. 
     
    \item \textbf{Food101}: The UPMC Food101 dataset \cite{Wang2015RecipeRW} contains web pages with textual recipe descriptions for 101 food. Each page is matched with a single image, which is obtained by querying Google Image Search for the given category. Typical examples of food labels include \texttt{Filet Mignon}, \texttt{Pad Thai}, \texttt{Breakfast Burrito}, and \texttt{Spaghetti Bolognese}. 
\end{itemize}

\begin{table*}
\caption{Comparison of our method with other competing approaches on the datasets of MMaterials and MAAPD.}
\label{table 2}
\centering
\setlength{\tabcolsep}{2.7mm}
\renewcommand{\arraystretch}{1.1}
\begin{tabular}{m{0.6cm}<{\centering}lcccccccc}
\toprule
~ & \multirow{2}{*}{Models} & \multicolumn{4}{c}{MMaterials} &  \multicolumn{4}{c}{MAAPD}\\
\cmidrule(lr){3-6}\cmidrule(lr){7-10}
~ & ~ & Accuracy & Precision & Recall & Macro-F1 & Accuracy & Precision & Recall & Macro-F1\\
\midrule
\multirow{12}{*}{\rotatebox{90}{\makecell{Single-modality\\baselines}}} & LR \cite{Pedregosa2011ScikitlearnML} & 84.6 & 84.7 & 83.0 & 83.3 & 79.9 & 76.8 & 74.5 & 75.4\\
~ & KimCNN \cite{kim2014convolutional} & 86.1 & 86.3 & 86.4 & 86.1 & 78.3 & 74.8 & 72.8 & 73.5\\
~ & FastText \cite{Joulin2017BagOT} & 86.5 & 87.0 & 85.2 & 85.8 & 80.2 & 76.8 & 74.7 & 75.3\\
~ & HAN \cite{yang2016hierarchical} & 88.4 & 88.3 & 88.6 & 88.4 & 76.9 & 72.9 & 72.7 & 72.3\\
~ & XML-CNN \cite{liu2017deep} & 88.1 & 86.8 & 89.3 & 87.8 & 79.7 & 76.2 & 75.1 & 75.3\\
~ & $\text{LSTM}_{reg}$ \cite{adhikari2019rethinking} & 87.1 & 86.8 & 86.8 & 86.6 & 78.5 & 74.0 & 72.7 & 73.0\\
\cmidrule(lr){2-10}
~ & RoBERT \cite{pappagari2019hierarchical} & 88.0 & 87.8 & 89.1 & 88.4 & 81.0 & 77.9 & 76.9 & 77.2\\
~ & ToBERT \cite{pappagari2019hierarchical}  & 87.9 & 87.5 & 89.1 & 88.2 & 81.8 & 78.7 & 77.4 & 77.8\\
~ & Longformer \cite{beltagy2020longformer}  & 88.9 & 88.1 & 89.9 & 88.7 & 81.5 & 80.3 & 76.0 & 77.4\\
~ & BigBird \cite{Zaheer2020BigBT} & 89.3 & 88.9 & 89.2 & 88.9 & 80.4 & 77.1 & 75.7 & 76.2\\
~ & Hi-Transformer\cite{Wu2021HiTransformerHI}  & 88.5 & 88.8 & 89.5 & 89.1 & 80.0 & 75.8 & 75.5 & 75.3\\
~ & HGCN \cite{Liu2022HierarchicalGC} & 89.6 & 88.6 & 90.1 & 89.3 & 79.9 & 76.8 & 75.0 & 75.7\\
\hline
\multirow{10}{*}{\rotatebox{90}{\makecell{Multi-modality\\baselines}}} & GMU \cite{Arevalo2017GatedMU} & 89.3 & 89.4 & 88.1 & 88.6 & 80.6 & 78.6 & 75.7 & 76.6\\
~ & TFN \cite{Zadeh2017TensorFN} & 89.9 & 89.5 & 89.6 & 89.4 & 79.7 & 79.2 & 73.5 & 74.7\\
~ & LMF \cite{Liu2018EfficientLM} & 89.1 & 89.1 & 89.1 & 89.0 & 80.8 & 78.1 & 74.5 & 75.7\\
~ & MISA \cite{Hazarika2020MISAMA} & 89.4 & 90.2 & 88.1 & 88.5 & 80.0 & 76.7 & 75.1 & 75.7 \\
~ & MMIM \cite{Han2021ImprovingMF} & 88.9 & 88.3 & 88.9 & 88.6 & 79.4 & 76.5 & 75.4 & 75.8\\
~ & ViLT \cite{Kim2021ViLTVT} & 87.9 & 89.0 & 87.5 & 87.6 & 78.8 & 73.8 & 70.4 & 71.5\\
~ & MMDynamics \cite{Han2022MultimodalDD} & 89.8 & 89.1 & 90.0 & 89.5 & 80.2 & 78.8 & 74.5 & 76.1\\
~ & UniS-MMC \cite{Zou2023UniSMMCMC} & 89.5 & 89.2 & 89.5 & 89.3 & 80.7 & 78.1 & 75.8 & 76.5\\
~ & DBF \cite{Wu2023DenoisingBW} & 89.6 & 90.7 & 88.9 & 89.4 & 82.4 & 79.8 & 78.0 & 78.5\\
~ & MetaPrompt \cite{zhao2024learning} & 89.2 & 89.7 & 88.2 & 88.3 & 80.3 & 78.0 & 75.5 & 76.4\\
\hline
~ & HMT & \textbf{90.8} & \textbf{90.6} & \textbf{91.5} & \textbf{90.9} & \textbf{83.8} & \textbf{82.0} & \textbf{79.4} & \textbf{80.3}\\
\bottomrule
\end{tabular}
\end{table*}

\subsection{Baselines and Metrics}
\subsubsection{Baselines}
We first compare the proposed HMT with several document classification methods, including Logistic Regression (LR) \cite{Pedregosa2011ScikitlearnML}, KimCNN \cite{kim2014convolutional}, FastText \cite{Joulin2017BagOT}, HAN \cite{yang2016hierarchical}, XML-CNN \cite{liu2017deep} and $\text{LSTM}_{reg}$ \cite{adhikari2019rethinking}, which can handle long texts and are provided by Hedwig\footnote{\url{http://hedwig.ca}}, a deep learning toolkit with pre-implemented document classification models, and multiple single-modality long document classification methods as follows,
\begin{itemize}
\setlength{\itemsep}{3pt}
\setlength{\parsep}{0pt}
\setlength{\parskip}{0pt}
    \item \textbf{RoBERT/ToBERT} \cite{pappagari2019hierarchical}: Splitting the input sequence into segments of a fixed size with overlap and then stacking these segment-level representations into a sequence, which serves as input to a small LSTM layer and a Transformer, respectively.
    
    \item \textbf{Longformer}\cite{beltagy2020longformer}: A variant of Transformer with local and global attention for long documents. 
    
    \item \textbf{BigBird}\cite{Zaheer2020BigBT}: The extending Longformer with another random attention.
    
    \item \textbf{Hi-Transformer}\cite{Wu2021HiTransformerHI}: Modeling long documents in a hierarchical manner and can capture the global document context for sentence modeling.

     \item \textbf{HGCN}\cite{Liu2022HierarchicalGC}:  Modeling long documents with a hierarchical graph convolutional network, in which a section graph and a word graph are constructed to model the macro and microstructure of long documents, respectively.
\end{itemize}

As for the multi-modal baselines, we select three related categories of methods, i.e. multi-modal classification \cite{Arevalo2017GatedMU, Zadeh2017TensorFN, Liu2018EfficientLM,Han2022MultimodalDD}, multi-modal emotion analysis \cite{Hazarika2020MISAMA, Han2021ImprovingMF}, and multi-modal transformer \cite{Kim2021ViLTVT}. For a fair comparison, we adopt the BigBird \cite{Zaheer2020BigBT}, which exhibits remarkable performance in the LDC field, as the text encoder of the multi-modal baselines. For images, except for the dataset of Review, which has provided the processed image features, we still adopt the CLIP model \cite{Radford2021LearningTV} for feature initialization. Additionally, the max-pooling operation will be performed on the image features if the multi-modal baselines are needed.
\begin{itemize}
\setlength{\itemsep}{3pt}
\setlength{\parsep}{0pt}
\setlength{\parskip}{0pt}
     \item \textbf{GMU} \cite{Arevalo2017GatedMU}: Receiving input from two or more sources and learning to decide how modalities influence the activation of the unit using multiplicative gates.

    \item \textbf{TFN} \cite{Zadeh2017TensorFN}: Modeling the dynamics of both intra- and inter-modality.
    
    \item \textbf{LMF} \cite{Liu2018EfficientLM}: Leveraging low-rank weight tensors to increase the effectiveness of multi-modal fusion without sacrificing performance.
    
    \item \textbf{MISA} \cite{Hazarika2020MISAMA}: Combining  modality-invariant and modality-specific features to predict emotional states. 
    
    \item \textbf{MMIM} \cite{Han2021ImprovingMF}: Maximizing mutual information in unimodal input pairings (inter-modality) and between the output of multi-modal fusion and unimodal input in a hierarchical manner.

     \item \textbf{ViLT} \cite{Kim2021ViLTVT}: Stemming from Vision Transformers \cite{Dosovitskiy2021AnII} and advancing to process multi-modal inputs with the tokenized texts and patched images.
     
    \item \textbf{MMDynamics} \cite{Han2022MultimodalDD}:  Dynamically evaluating both the feature-level and modality-level informativeness for different samples and thus trustworthily integrating multiple modalities.

    \item \textbf{UniS-MMC} \cite{Zou2023UniSMMCMC}: A novel method for multimodal contrastive learning aimed at deriving more reliable multimodal representations through the weak supervision provided by unimodal predictions.

    \item \textbf{DBF} \cite{Wu2023DenoisingBW}: A denoising bottleneck fusion model designed to proficiently manage extended longer multimodal sequences, effectively dealing with increased redundancy and noise.

    \item \textbf{MetaPrompt} \cite{zhao2024learning}: An innovative prompt learning paradigm featuring a dual-modality prompt tuning network that creates domain-invariant prompts, enabling smooth application across various multimodal domains.

\end{itemize}

\subsubsection{Metrics}
To comprehensively assess the performance of various methods, we employ four commonly used metrics: Accuracy, Precision, Recall, and F1 score in our experiments. These metrics provide valuable insights into the classification results. Let $TP, FP, TN, FN$ denote true positive, false positive, true negative, and false negative, respectively, and denote the total number of the samples as $N_{a}$. These metrics are defined as follows. 
\begin{equation}
    \text{Accuracy} = \frac{TP + TN}{N_{a}},  \text{Precision} = \frac{TP}{TP + FP},
\end{equation}
\begin{equation}
  \text{Recall} = \frac{TP}{TP + FN}, \text{F1-score} = \frac{2\times \text{Precision}\times \text{Recall}}{\text{Precision}+\text{Recall}}\label{fomulate 22}
\end{equation}
Here, we adopt the Macro-F1 metric by first calculating the f1-score for each label and then finding their unweighted mean as the final score.

\begin{table*}
\caption{Comparison of our method with other competing approaches on the datasets of Review and Food101.}
\label{table 3}
\centering
\setlength{\tabcolsep}{2.7mm}
\renewcommand{\arraystretch}{1.1}
\begin{tabular}{m{0.6cm}<{\centering}lcccccccc}
\toprule
~ & \multirow{2}{*}{Models} & \multicolumn{4}{c}{Review} &  \multicolumn{4}{c}{Food101}\\
\cmidrule(lr){3-6}\cmidrule(lr){7-10}
~ & ~ & Accuracy & Precision & Recall & Macro-F1 & Accuracy & Precision & Recall & Macro-F1\\
\midrule
\multirow{12}{*}{\rotatebox{90}{\makecell{Single-modality\\baselines}}} & LR \cite{Pedregosa2011ScikitlearnML} & 54.6 & 56.1 & 54.4 & 55.0 & 83.2 & 83.8 & 83.1 & 83.3\\
~ & KimCNN \cite{kim2014convolutional} & 56.3 & 55.7 & 56.1 & 55.6 &  88.8 & 89.6 & 88.7 & 88.9\\
~ & FastText \cite{Joulin2017BagOT} & 57.0 & 57.9 & 56.5 & 57.0 & 88.8 & 89.0 & 88.7 & 88.8\\
~ & HAN \cite{yang2016hierarchical} & 55.9 & 56.6 & 56.0 & 56.2 & 91.6 & 91.9 & 91.5 & 91.6\\
~ & XML-CNN \cite{liu2017deep} & 52.6 & 55.7 & 52.2 & 52.8 & 90.4 & 90.6 & 90.3 & 90.4\\
~ & $\text{LSTM}_{reg}$ \cite{adhikari2019rethinking} & 58.0 & 59.5 & 57.4 & 58.1 & 86.9 & 87.6 & 86.7 & 86.9\\
\cmidrule(lr){2-10}
~ & RoBERT \cite{pappagari2019hierarchical} & 58.4 & 57.7 & 58.6 & 57.5 & 90.8 & 92.7 & 90.7 & 91.4\\
~ & ToBERT \cite{pappagari2019hierarchical}  & 57.6 & 56.8 & 57.8 & 56.8 & 90.9 & 92.6 & 90.8 & 91.5\\
~ & Longformer \cite{beltagy2020longformer}  & 65.3 & 66.4 & 64.5 & 65.1 & 92.7 & 92.8 & 92.6 & 92.7\\
~ & BigBird \cite{Zaheer2020BigBT} & 64.3 & 65.4 & 64.1 & 64.5 &  92.7 & 92.8 & 92.6 & 92.7\\
~ & Hi-Transformer\cite{Wu2021HiTransformerHI}  & 57.3 & 59.1 & 56.2 & 57.0 & 90.4 & 90.5 & 90.3 & 90.4\\
~ & HGCN\cite{Liu2022HierarchicalGC}  & 67.8 & 68.5 & 67.2 & 67.7 & 91.5 & 91.7 & 91.4 & 91.5\\
\hline
\multirow{10}{*}{\rotatebox{90}{\makecell{Multi-modality\\baselines}}} & GMU \cite{Arevalo2017GatedMU} & 67.4 & 69.1 & 66.8 & 67.5 & 96.0 & 96.0 & 96.0 & 95.9\\
~ & TFN \cite{Zadeh2017TensorFN} & 65.9 & 66.0 & 65.7 & 65.8 & 93.5 & 93.5 & 93.4 & 93.5\\
~ & LMF \cite{Liu2018EfficientLM} & 63.8 & 64.5 & 63.4 & 63.8 & 93.2 & 93.2 & 93.2 & 93.2\\
~ & MISA \cite{Hazarika2020MISAMA} & 67.2 & 68.4 & 66.5 & 67.2 & 95.3 & 95.3 & 95.3 & 95.3\\
~ & MMIM \cite{Han2021ImprovingMF} & 63.2 & 64.2 & 63.2 & 63.5 & 94.7 & 94.7 & 94.7 & 94.7\\
~ & ViLT \cite{Kim2021ViLTVT} & 66.9 & 67.5 & 66.8 & 66.9 & 90.5 & 90.6 & 90.4 & 90.4\\
~ & MMDynamics \cite{Han2022MultimodalDD} & 65.3 & 66.6 & 64.4 & 65.2 & 95.7 & 95.7 & 95.7 & 95.7\\
~ & UniS-MMC \cite{Zou2023UniSMMCMC} & 68.6 & 68.7 & 68.7 & 68.7 & 95.4 & 95.4 & 95.4 & 95.4\\
~ & DBF \cite{Wu2023DenoisingBW} & 66.0 & 67.4 & 65.6 & 66.2 & 95.8 & 95.8 & 95.8 & 95.8\\
~ & MetaPrompt \cite{zhao2024learning} & 65.9 & 67.5 & 65.6 & 66.3 & 95.3 & 95.3 & 95.3 & 95.3\\
\hline
~ & HMT & \textbf{68.9} & \textbf{69.8} & \textbf{68.0} & \textbf{68.6} & \textbf{96.1} & \textbf{96.1} & \textbf{96.1} & \textbf{96.1}\\
\bottomrule
\end{tabular}
\end{table*}

\subsection{Implementation Details}
The max section length $r$, the total number of sentences, sections, and images of each sample are set to [256,100,8,9], [256,100,8,7], [512,25,1,5], [512,100,4,1] for MMaterials, MAAPD, Review, and Food101, respectively. The two multi-modal transformers are only superposed one layer in the actual employment, and the multi-scale window masks are set to [3,5,7,9,11] for MAAPD, and [3,5,7] for the other three datasets. The BERT-base-uncased and ViT-B/32 are selected as the pre-trained weights for the BERT and CLIP models, respectively. The models are trained for 30 epochs with a learning rate of 2e-5 and a weight decay of 0.1. Here, the parameters are updated with the AdamW optimizer with a mini-batch size of 4. Accuracy, Precision, Recall, and Macro-F1 are used as the evaluation metrics. Early stopping is performed when the F1 score of the validation set does not decrease for five consecutive epochs. We conduct all the experiments on an Nvidia 3090 GPU using the PyTorch deep learning framework. 

\subsection{Results and Analysis}
\subsubsection{Performance Comparison} 
Table~\ref{table 2} and Table~\ref{table 3} present the classification performance of HMT and the comparison methods on the four datasets in terms of Accuracy, Precision, Recall, and Macro-F1 score. Our approach significantly outperforms its counterparts, confirming its effectiveness in processing cross-modal long documents. A detailed analysis of the results is given in the following aspects.

\textbf{$\bullet$ Multi-modal baselines vs. Single-modal baselines:} Generally, due to more available data, multi-modal approaches typically outperform single-modality methods in performance. However, as shown in Table~\ref{table 2} and Table~\ref{table 3}, some multi-modal approaches, which process more information than single-modality methods, turn out to perform worse than the best outcomes of textual methods, contrary to what we have expected. This can be explained from two aspects. One reason is that the images contained in long documents usually contain much useless information and outliers. If not treated properly, it will have a certain impact on the final classification. Another reason is that the images in long documents have complex corresponding relationships with the texts, while this trait is not properly modeled by the typical multi-modal classification methods, which also has a certain impact on the final classification.

\textbf{$\bullet$ HMT vs. Single-modal baselines:} We can see from the results in Table~\ref{table 2} and Table~\ref{table 3} that our model significantly outperforms the single-modality baselines across all metrics. Due to the limited ability to capture long-range dependencies, the traditional document classification methods generally have poor classification performance except for HAN, which models long documents in a hierarchical manner. This demonstrates how crucial it is to represent structured features for long documents. Similarly, RoBERT, ToBERT,  Hi-Transformer, and HGCN also adopt hierarchical modeling methods and achieve certain performance improvements compared with HAN, mainly benefiting from their special input forms and information interaction modes. Owing to the sparse attention mechanism and the pre-trained operation on large-scale long document corpus, the Longformer and BigBird achieve further improvements on MAAPD and Food101. Our method capitalizes on the benefits of hierarchical text features and extends them by incorporating image features. The results show that our HMT model consistently outperforms the state-of-the-art methods on all datasets, validating the effectiveness of HMT in processing cross-modal long document data.

\begin{table*}
\caption{Architecture ablation study of HMT. Acc: Accuracy, Prec: Precision, Rec: Recall, F1: Macro-F1 Score. Best values are in bold}
\label{table 5}
\centering
\setlength{\tabcolsep}{1.6mm}
\renewcommand{\arraystretch}{1.2}
\begin{tabular}{ccccccccccccccccccc}
\toprule
\multicolumn{3}{c}{HMT} & \multicolumn{4}{c}{MMaterials} &  \multicolumn{4}{c}{MAAPD} & \multicolumn{4}{c}{Review} & \multicolumn{4}{c}{Food101}\\
\cmidrule(lr){1-3}\cmidrule(lr){4-7}\cmidrule(lr){8-11}\cmidrule(lr){12-15}\cmidrule(lr){16-19}
MMT & DMMT & DMT & Acc & Prec & Rec & F1 & Acc & Prec & Rec & F1 & Acc & Prec & Recall & F1 & Acc & Prec & Recall & F1\\
\midrule
~ & ~ & ~ & 89.1 & 88.2 & 90.4 & 89.1 & 80.6 & 77.8 & 76.7 & 76.9 & 65.9 & 67.7 & 65.4 & 66.1 & 92.1 & 92.2 & 92.0 & 92.1\\
\checkmark & ~ & ~ & 90.3 & 89.9 & 91.0 & 90.3 & 83.0 & 80.2 & 78.9 & 79.3 & 67.2 & 68.7 & 67.0 & 67.6 & 95.7 & 95.8 & 95.7 & 95.7\\
\checkmark & \checkmark & ~  & 90.4 & 90.4 & 91.4 & 90.7 & 83.0 & 79.6 & 79.3 & 79.4 & 68.0 & 68.7 & 67.6 & 67.9 & 96.1 & 96.1 & 96.1 & 96.1\\
\checkmark & \checkmark & \checkmark & \textbf{90.8} & \textbf{90.6} & \textbf{91.5} & \textbf{90.9} & \textbf{83.8} & \textbf{82.0} & \textbf{79.4} & \textbf{80.3} & \textbf{68.9} & \textbf{69.8} & \textbf{68.0} & \textbf{68.6} & \textbf{96.1} & \textbf{96.1} & \textbf{96.1} & \textbf{96.1}\\
\bottomrule
\end{tabular}
\end{table*}

\begin{figure*}[t]
  \centering
\includegraphics[width=\textwidth]{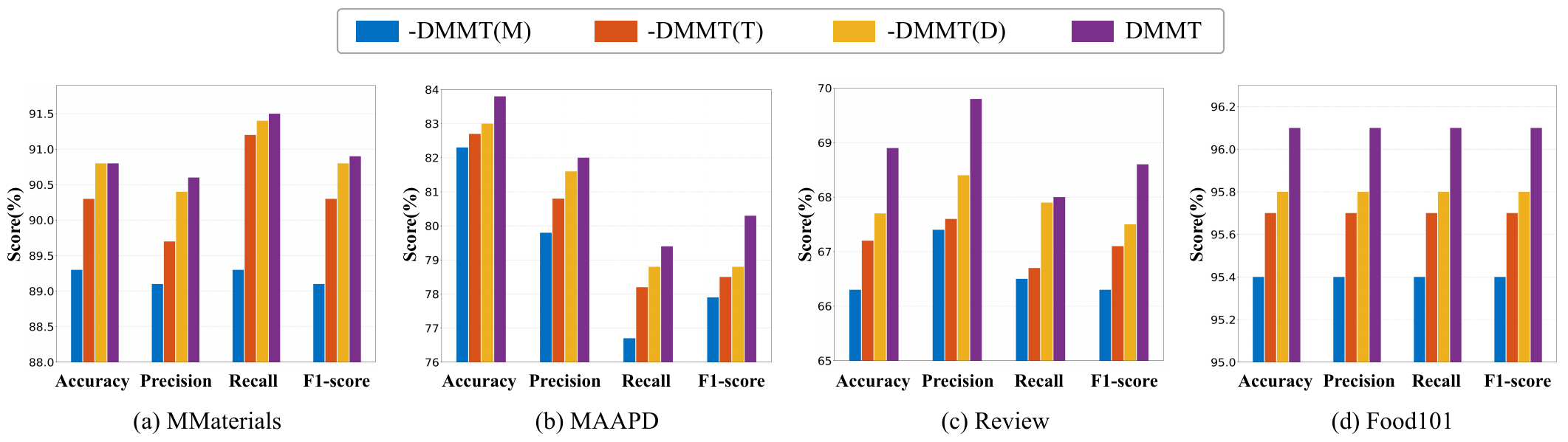}
  \caption{Effectiveness analysis of the DMMT module. -DMMT(M) denotes removing the multi-scale window masks. -DMMT(T) denotes removing the text branch of DMMT. -DMMT(D) denotes removing the dynamic fusion strategy and directly assigning the same weight to all branches.}
  \label{fig 10}
\end{figure*}

\textbf{$\bullet$ HMT vs. Multi-modal baselines:}
From the results in Table~\ref{table 2} and Table~\ref{table 3}, we can find that GMU, MMDynamics and UniS-MMC achieve better results than TFN and LMF, which indicates that simple fusion is preferable to complex fusion for cross-modal data without strict correspondence. The methods of MISA and MMIM focus on exploring mutual information or contrastive learning to narrow the feature distribution of the multi-modal features, but the existence of uncorrelated images will limit the performance of these methods. The Vision-Language Transformer, i.e., ViLT, which takes visual and textual embedding sequences as input, is more appropriate for processing short text-image pairs with strict correspondence. For the cross-modal long document with thousands of tokens and multiple images, the final performance can be easily impacted by the significant amount of noise information introduced during multi-modal interactions. DBF demonstrates strong performance across various datasets by effectively managing non-alignment, image redundancy, and data noise, making it suitable for multimodal long-document data. However, similar to other multimodal methods, DBF struggles to capture the structured information and multiscale correlations inherent in such data, despite its overall effectiveness. In contrast to methods focusing on single-scale text and image integration, our hierarchical multi-modal transformer (HMT) models structured information and multi-scale correlations in long documents. By employing multi-modal transformers at both section and sentence levels, our approach effectively captures complex relationships between text and image information. Experimental results consistently show superior classification performance across all datasets, confirming the effectiveness of our model. 

\subsection{Ablation Study} 
An ablation study is conducted on the four datasets to examine the effects of each component of HMT. The results are presented in Table~\ref{table 5} and Fig.~\ref{fig 10}. Specifically, we begin with the base model that utilizes only the section features. Then, we incorporate image features and use the Multi-modal Transformer (MMT) to model inter-modal correlations. Based on the results shown in Table~\ref{table 5}, we can observe that integrating image information using the multi-modal transformer has led to significant performance improvements across all metrics on the four datasets. This indicates the benefits of incorporating image features into the model. Furthermore, by introducing the Dynamic Multi-scale Multi-modal Transformer (DMMT) to model the complex correlations between sentences and images, we achieve even better results compared to the former model. This highlights the importance of capturing multi-modal interactions at the sentence level to extract more discriminative features. Finally, by further incorporating the Dynamic Mask Transfer (DMT) module into the model to form the full architecture, we achieve the best results compared with other ablation models, which demonstrates the effectiveness of the hierarchical structure information, and simultaneously confirms the validity of our model in combining multiple images with the hierarchical text features of long documents.

In order to further demonstrate the impact of different parts of the DMMT module, as illustrated in Fig.~\ref{fig 10}, we present three variations of the DMMT module: -DMMT(M) represents the removal of multi-scale window masks; -DMMT(T) represents the removal of the text branch of DMMT; and -DMMT(D) represents the removal of the dynamic weight generation and assigns equal weights to all branches. The results show a significant decrease in performance when the multi-scale window masks are removed, indicating that modeling the multi-scale correspondence of sentence and image features is crucial for obtaining a more discriminative representation of long documents. Comparing the performance of DMMT with that of -DMMT(T) shows that the text branch in DMMT can effectively mitigate the influence of irrelevant images on the final representation of long documents. Moreover, the lack of the dynamic fusion mechanism also leads to performance degradation, demonstrating the significant role played by the dynamic fusion mechanism in achieving substantial improvements.

\begin{table}
\caption{Performance comparison of Macro-F1 Score under window masks with different numbers and scales.}
\label{table 4}
\centering
\setlength{\tabcolsep}{3.1mm}
\renewcommand{\arraystretch}{1.3}
\begin{tabular}{lcccc}
\toprule
Models & MMaterials &  MAAPD & Review & Food101\\
\midrule
3 & 88.8 & 78.2 & 66.4 & 95.4\\
3,5 & 89.6 & 78.3 & 67.5 & 96.0\\
3,5,7 & \textbf{90.9} & 78.9 & \textbf{68.6} & \textbf{96.1}\\
3,5,7,9 & 89.9 & 79.7 & 67.0 & 95.6\\
3,5,7,9,11 & 89.2 & \textbf{80.3} & 66.1 & 95.4\\
\bottomrule
\end{tabular}
\end{table}

\subsection{Hyperparameter Study}
Section length $r$, the number of HMT layers $N$, and the number of multi-scale window masks $n_{win}$ are the three major hyperparameters of our HMT model. To find the best configuration, we do extensive CLDC experiments on the validation set using various values for these hyperparameters.

\begin{figure}
    \centering
\setlength{\abovecaptionskip}{0pt}
\setlength{\belowcaptionskip}{-6pt}
    \subfigure[Section Length]{\label{fig 6}\includegraphics[width=0.24\textwidth]{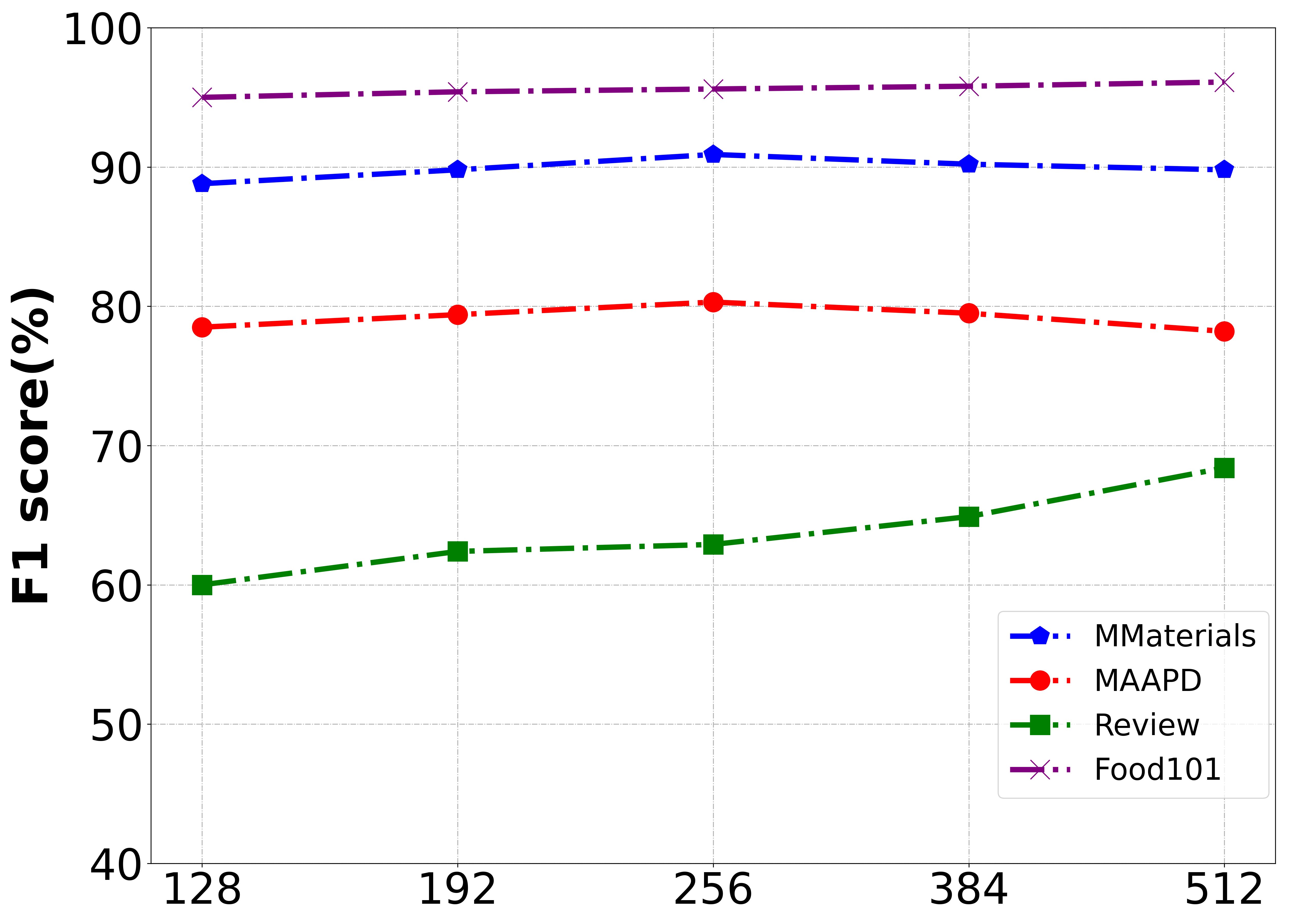}}
    \subfigure[Number of Layers]{\label{fig 7}\includegraphics[width=0.24\textwidth]{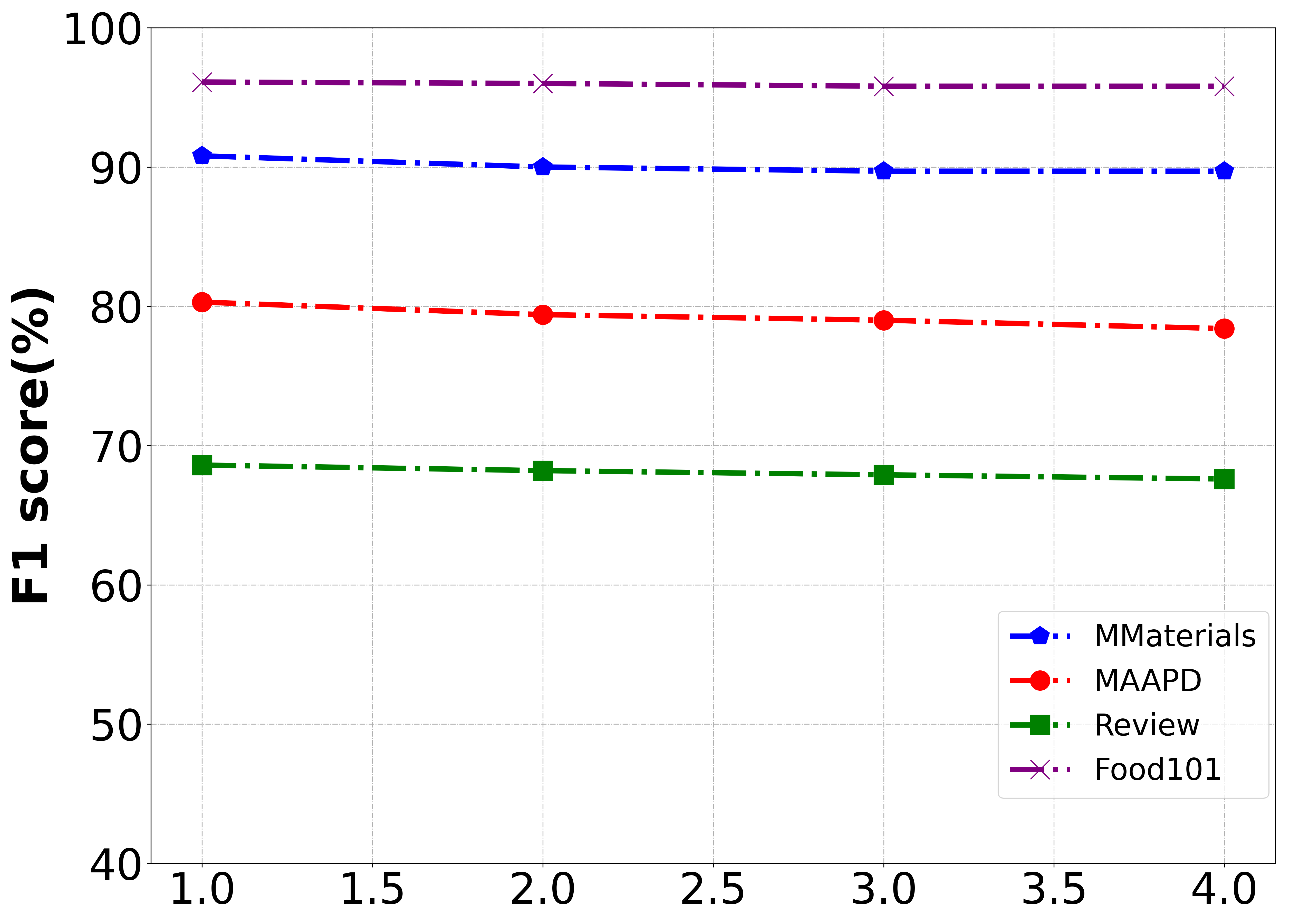}}
    \caption{Performance vs. Parameters.}
    \label{fig 9}
\end{figure}

\textbf{$\bullet$ Section Length $r$:} 
In our model, the size of the section length $r$ has a direct impact on the section and sentence representations, thus affecting subsequent cross-modal interactions. Limited by the maximum modeling length of pre-trained BERT, we select five different section lengths, i.e. 128, 192, 256, 384, 512 for each dataset and plot the F1-score as a function of $r$ in Fig.~\ref{fig 6}.  Intuitively, a shorter section length means more accurate text encoding. However, it may not always have a high degree of compatibility with image features. For MMaterials and MAAPD with multiple images, we find that the best results can be obtained when the section length is set to 256. For the datasets of Review and Food101, the overall performance is enhanced when the section length increases from 128 to 512. This can be explained that the image information of these two datasets is more skewed toward the expression of overall information, and to some extent, the more macroscopic text features will be semantically related to the image features.

\textbf{$\bullet$ Number of HMT Layers $N$:} The influence of the layer numbers is illustrated in Fig.~\ref{fig 7}. We can see the decrease in our model performance when the number of layers increases from 1 to 4. We deem that this is primarily because both the initial text and image features are derived from the large-scale pre-trained models and classified as high-level semantic features. With the powerful interaction capabilities of the Transformer, it is possible to achieve effective information interaction with fewer layers. Additionally, the introduction of the multi-scale transformer further strengthens the information interaction ability of the single layer.

\textbf{$\bullet$ Number of Window Masks $N_{win}$:} Table~\ref{table 4} shows the results of the model using window masks with different numbers and scales. From the results, we can see that the best results are obtained when the multi-scale window masks are set to [3,5,7] for MMaterials, Review, and Food101. 
This suggests that increasing the number of multi-scale window masks does not always lead to better results. The plethora of multi-scale interactions will introduce additional noise information, limiting the final performance of the model. For the dataset of MAAPD, we find that the model performs best when the multi-scale window masks are set to [3,5,7,9,11], implying that the overall association between text features in these scales and corresponding image features is higher.

\begin{table}
\caption{Computational complexity on time (s/epoch) and space (Mb) on the dataset of MMaterials.}
\label{table 6}
\centering
\setlength{\tabcolsep}{1.0mm}
\renewcommand{\arraystretch}{1.2}
\begin{tabular}{lccc}
\toprule
\multirow{2}{*}{Models} &
\multicolumn{3}{c}{MMaterials}\\
\cmidrule(lr){2-4}
~ & Training(s/epoch) & Inference(s/epoch) & \#Memory(Mb)\\
\midrule
\multicolumn{4}{l}{\emph{Single-modality baselines}}\\
\hline
RoBERT \cite{pappagari2019hierarchical} & 1264 & 72 & 18613\\
ToBERT \cite{pappagari2019hierarchical} & 1270 & 71 & 18721\\
Longformer \cite{beltagy2020longformer} & 1417 & 59 & 21871\\
BigBird \cite{Zaheer2020BigBT} & 1275 & 53 & 23707\\
Hi-Transformer \cite{Wu2021HiTransformerHI} & 750 & 42 & 18075\\
HGCN \cite{Liu2022HierarchicalGC} & 571 & 35 & 17553\\
\hline
\multicolumn{4}{l}{\emph{Multi-modality baselines}}\\
\hline
GMU \cite{Arevalo2017GatedMU}  & 1307 & 55 & 23861\\
TFN \cite{Zadeh2017TensorFN} & 1314 & 54 & 23857\\
LMF \cite{Liu2018EfficientLM} & 1322 & 55 & 23845\\
MISA \cite{Hazarika2020MISAMA}  & 1310 & 55 & 23719\\
MMIM \cite{Han2021ImprovingMF} & 1281 & 54 & 23839\\
ViLT \cite{Kim2021ViLTVT} & 1208 & 66 & 29249\\
MMDynamics \cite{Han2022MultimodalDD} & 1304 & 54 & 23737\\
UniS-MMC \cite{Zou2023UniSMMCMC} & 1505 & 65 & 20575\\
DBF \cite{Wu2023DenoisingBW} & 881 & 47 & 13551\\
MetaPrompt \cite{zhao2024learning} & 936 & 49 & 18989\\
\hline
HMT & 967 & 53 & 12537\\
\hline
\end{tabular}
\end{table}

\subsection{Computational Complexity}
In this part, we present a comparison of the computational complexity of our proposed method with other baselines, except for traditional document classification methods that show relatively poor performance on the four datasets. Table~\ref{table 6} summarizes the results, which indicate that our method has certain disadvantages in terms of time and space complexity compared to the single-modality LDC methods Hi-Transformer and HGCN. This is mostly due to the comparatively simple structures of these two models and the addition of pooling operations further expanding the advantages of the HGCN model. However, our method achieves significant performance improvements compared to these two methods. In addition, compared to other single-modality and multi-modality baselines, our method demonstrates smaller space complexity and competitive time complexity, validating that our model can achieve a favorable speed-accuracy trade-off.

\subsection{Visualization}
\begin{figure}
    \centering
\setlength{\abovecaptionskip}{0pt}
\setlength{\belowcaptionskip}{-6pt}
    \subfigure{\label{fig 11}\includegraphics[width=0.5\textwidth]{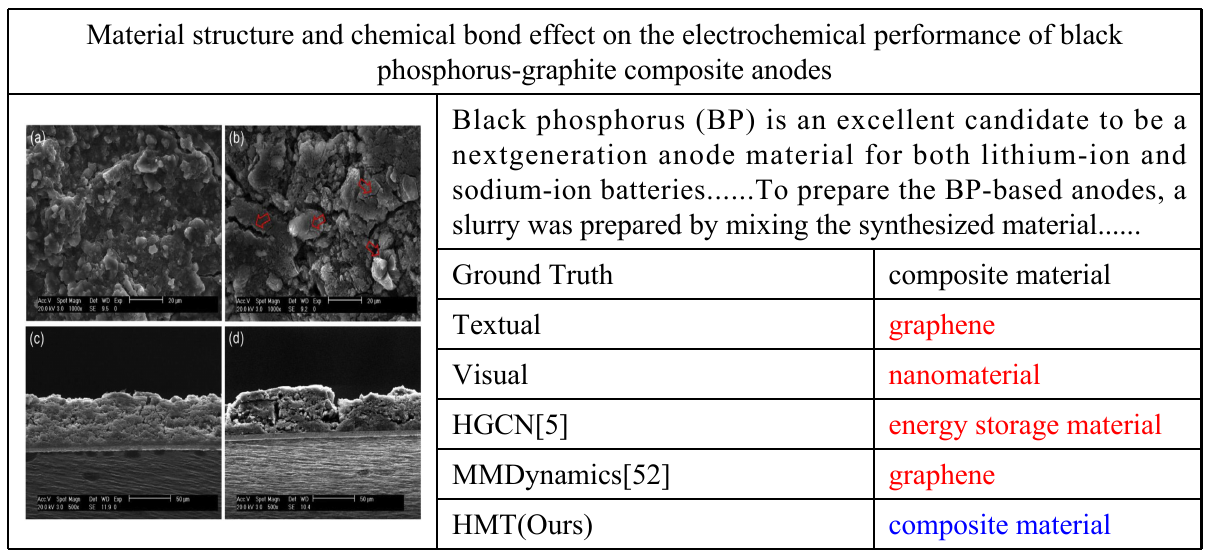}}
    \subfigure{\label{fig 12}\includegraphics[width=0.5\textwidth]{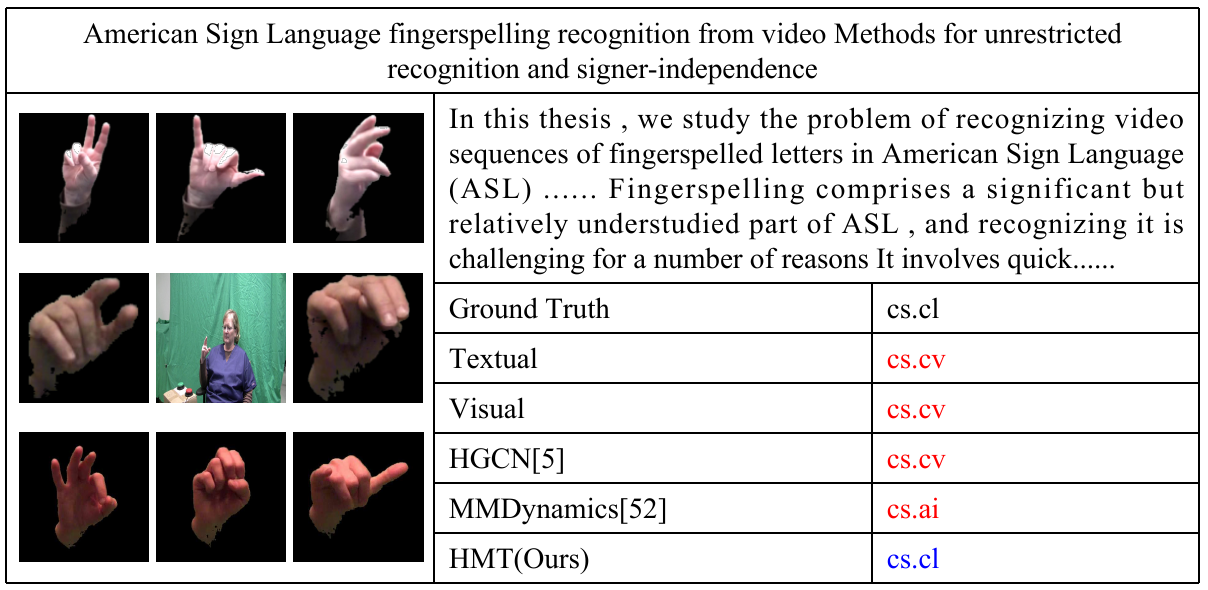}}
    \caption{Examples of predictions in test set. Red and blue genres are false positives and true positives, respectively.}
    \label{fig 13}
\end{figure}

To better understand the efficacy of our proposed model, we visualize some examples of the datasets MMaterials and MAAPD and provide the outcomes of their classification under different model inputs. 
From the qualitative results shown in Fig.~\ref{fig 13}, we can find that both test samples produce inaccurate classification results when employing single-modality information, implying that the discrimination capability of single-modality information is subpar in some situations. 
Furthermore, we included a comparison with the high-performing single-modal method HGCN \cite{Liu2022HierarchicalGC} and the multi-modal method MMDynamics \cite{Han2022MultimodalDD}. Interestingly, both methods produced incorrect classification results on these two samples. Our analysis indicates that the significant ambiguity in the textual content poses a challenge for purely text-based methods to accurately interpret its meaning. Conversely, the multi-modal method MMDynamics, which focuses on fusing overall textual and visual features, does not sufficiently capture the multi-scale correspondences between these modalities. Conversely, our model achieves accurate predictions by incorporating hierarchical multi-scale modal interactions, demonstrating the complementarity between multi-modal information and confirming the efficacy of our proposed method in handling cross-modal long document data.

\section{Conclusion}\label{Conclusion}
In this paper, we propose a novel hierarchical multi-modal transformer for cross-modal long document classification, which performs multi-modal interaction and fusion at both the section and sentence levels. Concretely, two multi-modal transformers are implemented respectively to model the complex correspondence between the image features and the section and sentence features. Additionally, to fully utilize the hierarchical structure information, a dynamic mask transfer module is further introduced to integrate the two multi-modal transformers as a whole by propagating features between them. Extensive experiments conducted on two newly created and two publicly available multi-modal long document datasets demonstrate that our proposed HMT can achieve significant and consistent improvements on the CLDC task. Although the hierarchical text features of long documents are represented in our model, the structure of the images is not deeply discussed, which is a direction of our future
work.

\bibliographystyle{IEEEtran}
\bibliography{HMT}

\vfill

\end{document}